\documentclass[conference]{IEEEtran}

\usepackage[utf8]{inputenc}
\usepackage{cite}
\usepackage{graphicx}
\usepackage{amsmath,amssymb,amsfonts}
\usepackage{algorithmic}
\usepackage{textcomp}
\usepackage{xcolor}
\usepackage{url}
\usepackage{array}    
\usepackage{listings} 
\usepackage{seqsplit}
\usepackage{capt-of} 
\usepackage{placeins} 

\usepackage{filecontents}
\usepackage{mdframed} 

\usepackage{booktabs}
\usepackage{longtable}
\usepackage{array}
\usepackage{geometry}
\usepackage{ragged2e} 

\newcolumntype{P}[1]{>{\RaggedRight\arraybackslash}p{#1}}
\definecolor{light-gray}{gray}{0.95}

\def\BibTeX{{\rm B\kern-.05em{\sc i\kern-.025em b}\kern-.08em
    T\kern-.1667em\lower.7ex\hbox{E}\kern-.125emX}}

\begin{document}

\title{Simpliflow: A Lightweight Open-Source Framework for Rapid Creation and Deployment of Generative Agentic AI Workflows}

\author{\IEEEauthorblockN{Deven Panchal}
\IEEEauthorblockA{\textit{Senior Member, IEEE} \\
\textit{Georgia Institute of Technology}\\
USA\\
Email: devenrpanchal@gatech.edu}
}

\maketitle

\begin{abstract}
Generative Agentic AI systems are emerging as a powerful paradigm for automating complex, multi-step tasks. However, many existing frameworks for building these systems introduce significant complexity, a steep learning curve, and substantial boilerplate code, hindering rapid prototyping and deployment. This paper introduces simpliflow, a lightweight, open-source Python framework designed to address these challenges. simpliflow enables the rapid development and orchestration of linear, deterministic agentic workflows through a declarative, JSON-based configuration. Its modular architecture decouples agent management, workflow execution, and post-processing, promoting ease of use and extensibility. By integrating with LiteLLM, it supports over 100 Large Language Models (LLMs) out-of-the-box. We present the architecture, operational flow, and core features of simpliflow, demonstrating its utility through diverse use cases ranging from software development simulation to real-time system interaction. A comparative analysis with prominent frameworks like LangChain and AutoGen highlights simpliflow's unique position as a tool optimized for simplicity, control, and speed in deterministic workflow environments.
\end{abstract}


\section{Introduction}

\begin{figure*}[!t]
    \centering
    \includegraphics[width=\textwidth]{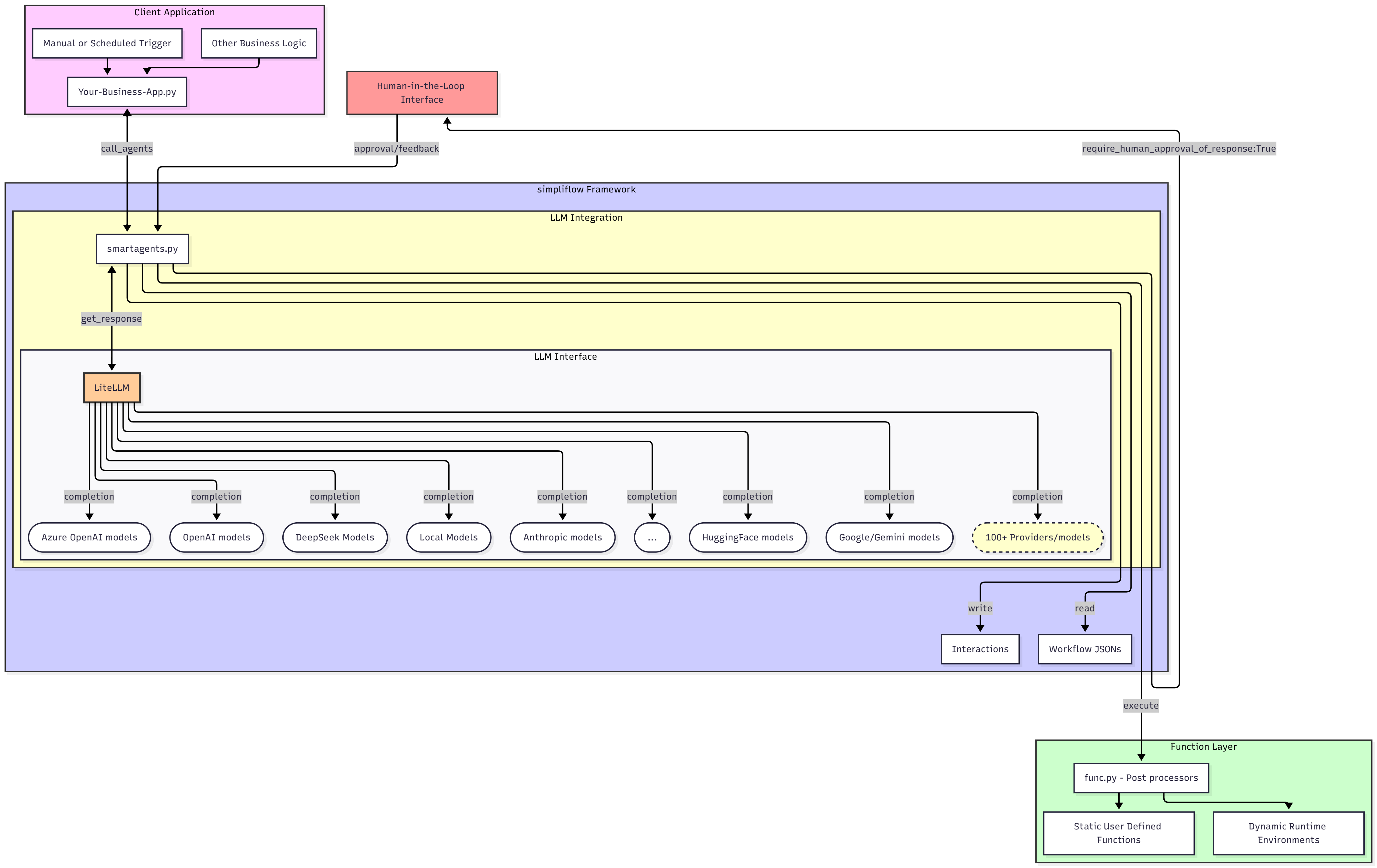}
    \caption{High-level System Architecture of the simpliflow framework, illustrating the interaction between the client application, the LLM integration layer, the human-in-the-loop interface, and the function layer.}
    \label{fig:sys_arch}
\end{figure*}

Agentic AI refers to AI systems composed of autonomous or semi-autonomous “agents” that can independently carry out tasks or collaborate in sequences to achieve goals. These agents are typically powered by large language models (LLMs) and can perform both simple and complex tasks with minimal human intervention. Such agentic AI workflows have enormous potential across domains – from automating business processes to generating creative content – offering efficiency gains and novel capabilities (e.g. significant cost savings or new product opportunities). However, orchestrating multiple AI agents and tools in a reliable way is challenging. It requires managing prompts, tool invocations, data passing between steps, and often human oversight to ensure correct and safe operation.

Existing agentic AI toolkits—such as LangChain \cite{langchain_github}, AutoGen \cite{autogen_github}, and BabyAGI \cite{babyagi_github}—unlock powerful capabilities (multi-agent collaboration, tool use, memory), but they often demand substantial setup and expertise and tend to execute in non-deterministic ways that complicate debugging, reproducibility, and onboarding for fast experimentation.

The primary contribution of this paper is the introduction of simpliflow \cite{simpliflow_github}. simpliflow is an Open-Source Modular Framework for Generative AI Agentic Workflow Orchestration. Distributed as a Python package (\texttt{pip install simpliflow}) it supports 100+ LLM models/vendors through LiteLLM \cite{litellm_github} and allows developers to spin up chains of LLM-powered agents interwoven with custom user-defined functions in seconds. Simpliflow dramatically reduces the time, effort, and specialized knowledge required to incorporate generative agentic AI into applications and workflows at scale. It runs on the IDE of your choice (VSCode, Pycharm, Spyder, Jupyter Notebooks, etc.) and on your platform/OS (Windows, Linux, Mac). By defining a workflow in a simple JSON configuration, users can orchestrate an infinitely long sequence of agent steps with minimal coding. This lowers the barrier to adoption for AI automation: even complex multi-step tasks can be automated or accelerated without building a bespoke agent system from scratch. It presents a lightweight alternative that trades some autonomy for clarity and control: it models workflows as deterministic, linear FSMs with a single, predictable transition per step, avoiding the confusing chat/agent semantics and API fragility reported in other frameworks. simpliflow's JSON-based workflow definition makes orchestration transparent and auditable, while the engine enforces stepwise sequencing, supports human-in-the-loop approvals, and logs every interaction as structured JSON for inspection. The framework's nature effortlessly allows AI by AI for AI paradigm i.e. design-time AI support to create AI that can do AI. AI (agents) by AI (in-IDE AI) to do AI (final task) can help you write and execute intelligent machine learning and AI programs using just a high-level intent in natural language i.e. English - all while controlling the creativity, factual nature and diversity of the outputs. It embraces pluggable postprocessor functions—from simple formatters and validators to “AI-to-Action” code execution—so developers can inject domain logic or trigger real actions without leaving the flow. This AI-to-Action feature (code execution) can actually execute the results of the agentic interactions in-situ to for e.g. convert your laptop to a music synthesizer and player or even create and execute quantum programs on a real remote Quantum Computer – all from within the framework.

In the following, we detail Simpliflow's system architecture and key features, usage, examples and compare it to other prominent agentic AI frameworks (LangChain \cite{langchain_github}, AutoGen \cite{autogen_github}, BabyAGI \cite{babyagi_github}, CrewAI \cite{crewai_github}, SuperAGI \cite{superagi_github}, etc.)

\section{System Architecture}
Simpliflow follows a modular architecture that cleanly separates the definition of an agent workflow from its execution engine. The core components of the framework include the Agent class, the Workflow Engine, Postprocessor functions, and a JSON-based configuration system for workflows:

\begin{itemize}
    \item \textbf{Agent Class}: Simpliflow provides an Agent class (in \texttt{smartagents.py}) which serves as a blueprint for creating agents. Each agent represents a single step in the workflow. An agent is defined by attributes such as its name, role, and task description (prompt), and it encapsulates the logic to communicate with an LLM. Internally, the Agent class leverages LiteLLM \cite{litellm_github} – a lightweight library – to interface with various LLMs (OpenAI GPT series, Anthropic Claude, Google models, etc.) behind a common API. This design abstracts away the specifics of different model providers, allowing multi-vendor and multi-model integration transparently. When executed, an agent invokes an LLM with its prompt, producing an output.
    
    \item \textbf{simpliflow Workflow Engine/ simpliflow.smartagents}: The engine orchestrates the sequence of agent executions and data flow between them. Workflows are defined as directed sequences of agents (forming a linked list or finite state machine of sorts), and the engine ensures each agent's output is passed as input to the next agent in the chain. The engine also handles interaction logging and human-in-the-loop (HITL) processes. For every run, it creates an interaction log (JSON file) recording each agent's input prompt and output. The interactions are stored in the Interactions directory, with filenames derived from the workflow file (e.g., Simple-Quantum-Circuit-Creator-And-Executor\_interactions.json for the Simple-Quantum-Circuit-Creator-And-Executor.json workflow). If an agent is marked as requiring human approval, the engine will halt after that agent's output and await user confirmation before proceeding. This allows a human supervisor to validate or edit an intermediate result. The workflow engine thus acts as the deterministic "conductor,” advancing through the predefined steps and maintaining state (the current agent and any interim data) until completion or interruption. The overall system architecture is shown in Fig. \ref{fig:sys_arch}.
    
    \item \textbf{Postprocessor Functions}: In many workflows, it is useful to manipulate or validate the raw LLM output from an agent before passing it on. Simpliflow allows arbitrary \textbf{postprocessor functions} (written in Python) to be plugged in after any agent to modify its output. These functions can be written by the user in \texttt{func.py}, receive the agent's output and allow for tasks such as data extraction, format conversion, and validation. Basically, anything a user may want to do. The framework allows for dynamic injection of these functions into the workflows. Examples could be static user-defined functions like \texttt{pingserver}, \texttt{printinpink}, etc. Or they could be dynamic runtime environments/interpreters/utilities to actually execute code and scripts like \texttt{execute\_python\_code}. This capability allows simpliflow to not only generate intelligent output but \textit{take actions} based on that output – for instance, turning your laptop into a music synthesizer by executing generated audio code, or running a query on a real quantum computer from an LLM-generated program.
    
    \item \textbf{Configuration System: JSON-Based Workflow Definition}: Simpliflow emphasizes configuration-over-code for defining agentic workflows. A workflow is described in a single JSON file which lists all the agents (steps) and their properties, as well as how they connect. Each agent entry in the JSON has to be specified with its role, task, whether or not human approval is required for its output, any post-processing function, etc.
\end{itemize}

Example JSON workflow looks like below:

\begin{mdframed}[backgroundcolor=light-gray, roundcorner=10pt,leftmargin=1, rightmargin=1, innerleftmargin=2, innertopmargin=5,innerbottommargin=5, outerlinewidth=1, linecolor=light-gray]
\begin{lstlisting}[basicstyle=\scriptsize\ttfamily, breaklines=true]
{
  "flow_description": "Give the workflow some name",
  "agents": [
    {
      "head": "True",
      "name_of_agent": "Agent1",
      "role_of_agent": "First Agent",
      "what_should_agent_do": "Task description",
      "require_human_approval_of_response": "False",
      "postprocessor_function": "process_output",
      "next": "Agent2"
    },
    {
      "head": "False",
      "name_of_agent": "Agent2",
      "role_of_agent": "Second Agent",
      "what_should_agent_do": "Next task",
      "require_human_approval_of_response": "True",
      "postprocessor_function": "None",
      "next": "None"
    }
  ]
}
\end{lstlisting}
\end{mdframed} 

JSON schema to validate the workflow JSON files:

\begin{mdframed}[backgroundcolor=light-gray, roundcorner=10pt,leftmargin=1, rightmargin=1, innerleftmargin=2, innertopmargin=5,innerbottommargin=5, outerlinewidth=1, linecolor=light-gray]
\begin{lstlisting}[,basicstyle=\scriptsize\ttfamily, breaklines=true]
{
    "$schema": "http://json-schema.org/draft-07/schema#",
    "title": "Workflow Schema",
    "description": "Schema for validating workflow JSON files",
    "type": "object",
    "required": ["flow_description", "agents"],
    "properties": {
        "flow_description": {
            "type": "string",
            "description": "A description of what the workflow does"
        },
        "agents": {
            "type": "array",
            "description": "Array of agents that form the workflow",
            "minItems": 1,
            "items": {
                "type": "object",
                "required": [
                    "head",
                    "name_of_agent",
                    "role_of_agent",
                    "what_should_agent_do",
                    "require_human_approval_of_response",
                    "postprocessor_function",
                    "next"
                ],
                "properties": {
                    "head": {
                        "type": "string",
                        "enum": ["True", "False"],
                        "description": "Whether this agent is the first agent in the workflow"
                    },
                    "name_of_agent": {
                        "type": "string",
                        "description": "The name identifier for the agent"
                    },
                    "role_of_agent": {
                        "type": "string",
                        "description": "The role or purpose of the agent in the workflow"
                    },
                    "what_should_agent_do": {
                        "type": "string",
                        "description": "The task or instruction for the agent"
                    },
                    "require_human_approval_of_response": {
                        "type": "string",
                        "enum": ["True", "False"],
                        "description": "Whether human approval is required before proceeding"
                    },
                    "postprocessor_function": {
                        "type": "string",
                        "description": "Name of the postprocessor function to apply to the agent's output"
                    },
                    "next": {
                        "type": "string",
                        "description": "Name of the next agent in the workflow sequence or 'None' if this is the last agent"
                    }
                },
                "additionalProperties": false
            }
        }
    },
    "additionalProperties": false
}

\end{lstlisting}
\end{mdframed} 

\begin{figure}[!h]
    \centering
    \includegraphics[width=\columnwidth]{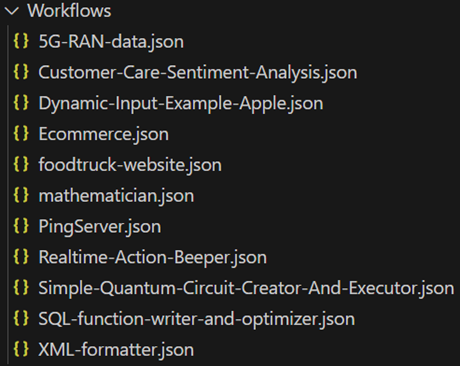}
    \caption{Example workflows provided in the "Workflows" directory of the simpliflow-usage repository \cite{simpliflow_usage_github}.}
    \label{fig:workflows_list}
\end{figure}

Github Repository simpliflow-usage \cite{simpliflow_usage_github}, comes with many readymade workflows from various domains. Users can run these or customize these or create their own workflows. A sample of these workflows is shown in Fig. \ref{fig:workflows_list}.

\begin{figure}[!h]
    \centering
    \includegraphics[width=\columnwidth]{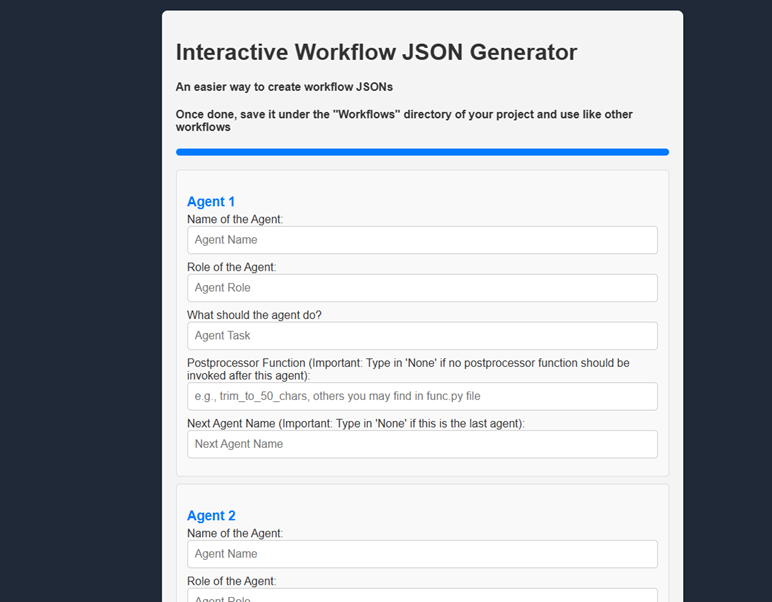}
    \caption{The Interactive Workflow JSON Generator web utility provided to easily create workflow configuration files.}
    \label{fig:interactive_generator}
\end{figure}

The workflows can be crafted by hand or using assistance from In-IDE AI's like in GitHub Copilot, Cursor etc. or using the Interactive Workflow JSON Generator utility (Fig. \ref{fig:interactive_generator}) that comes along with simpliflow in the simpliflow-usage repo \cite{simpliflow_usage_github}.

\section{Usage, Execution, Accessing Results and Visualization}
Users should first install the python package using \texttt{pip install simpliflow}.

\begin{figure*}[!t]
    \centering
    \includegraphics[width=\textwidth]{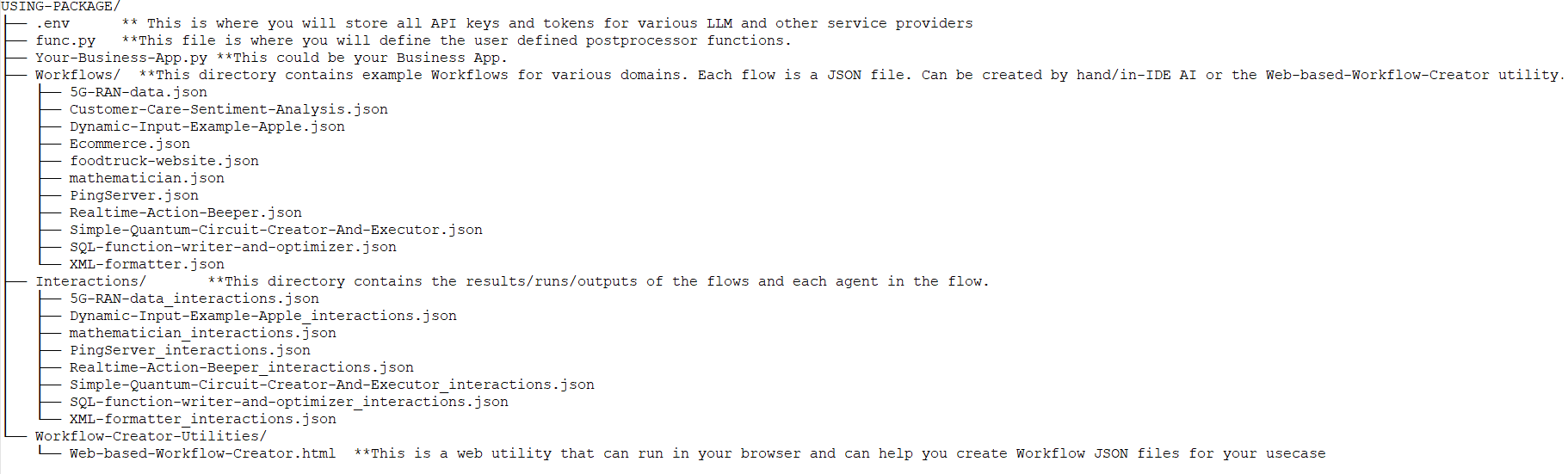}
        \caption{Directory set-up required when using simpliflow.}
    \label{fig:direct_struc}
\end{figure*}

\begin{figure}[!h]
    \centering
    \includegraphics[width=\columnwidth]{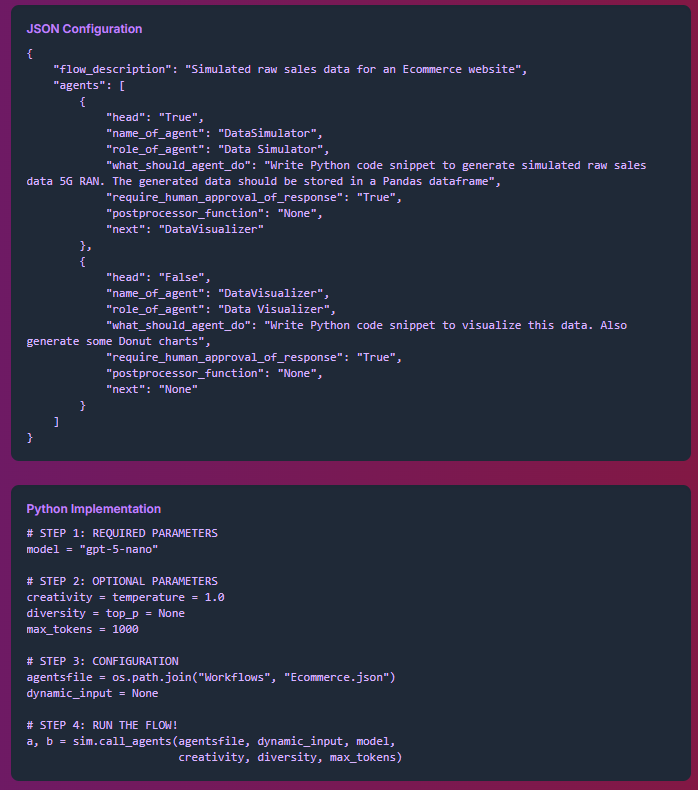}
    \caption{An example JSON workflow configuration (top) and the corresponding Python code in a business application to trigger its execution (bottom).}
    \label{fig:json_and_code}
\end{figure}

The users can set up a directory similar to the \texttt{USING-PACKAGE} directory that is also shown in \ref{fig:direct_struc}. This directory would ideally be any directory where you would hold the Workflows you define, its interactions, the API tokens in a \texttt{.env} file and also the script that will finally run the flows. In our case we name it \texttt{Your-Business-App.py}. The way to run a workflow is to set required parameters, set optional parameters, some more configuration about pointing the \texttt{agentsfile} to an appropriate workflow json, setting inputs/dynamic inputs for the workflow etc, and then finally running the workflow using \texttt{sim.call\_agents()} function from the simpliflow package, as demonstrated in Fig. \ref{fig:json_and_code}.

\begin{figure}[!h]
    \centering
    \includegraphics[width=\columnwidth]{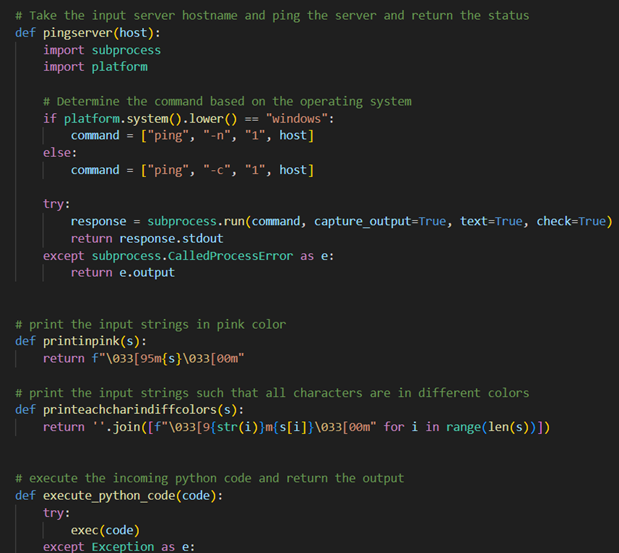}
    \caption{Example user-defined postprocessor functions in \texttt{func.py}, including network utilities, text formatters, and a Python code executor.}
    \label{fig:func_py}
\end{figure}

\begin{figure}[!h]
    \centering
    \includegraphics[width=\columnwidth]{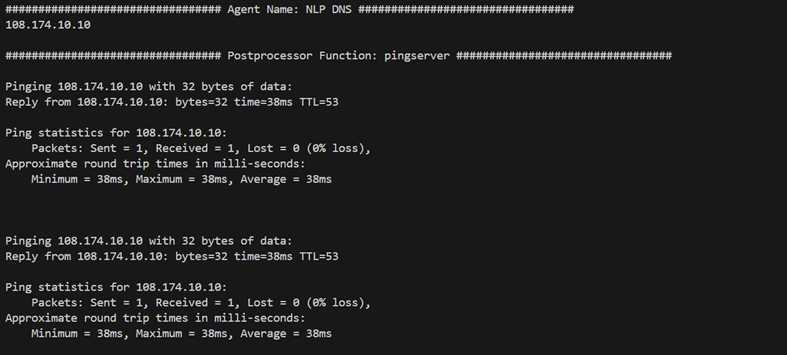}
    \caption{Terminal output from the execution of the "PingServer.json" workflow, showing the result of the \texttt{pingserver} postprocessor function.}
    \label{fig:pingserver_execution}
\end{figure}

\begin{figure}[!h]
    \centering
    \includegraphics[width=\columnwidth]{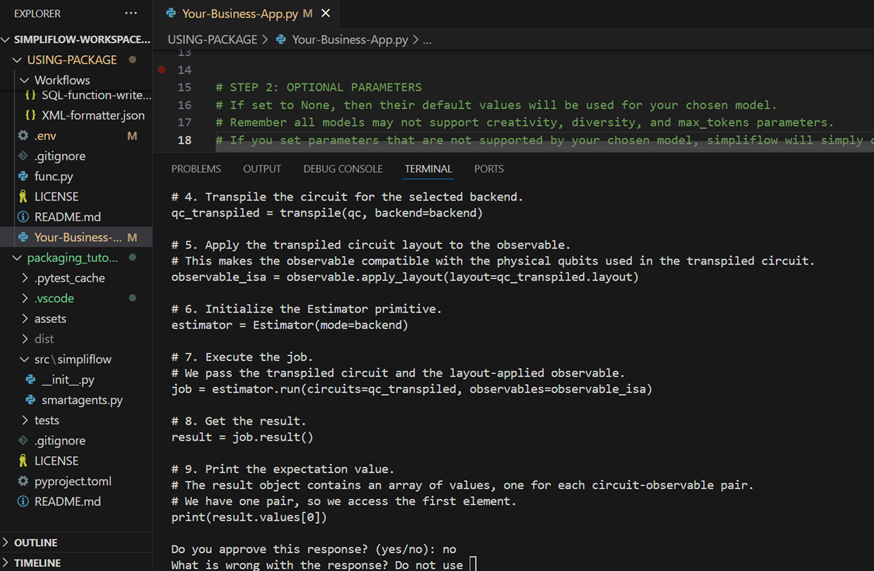}
    \caption{Example of the Human-in-the-Loop (HITL) feature, where the framework pauses execution and prompts for user approval in the terminal.}
    \label{fig:human_approval}
\end{figure}

There is support for User-defined Post-Processor functions and their configuration and execution. They can be defined in \texttt{func.py} and used in the workflow JSONs (Fig. \ref{fig:func_py}). Fig. \ref{fig:pingserver_execution} shows the terminal output when a workflow with the \texttt{pingserver} postprocessor function is executed. Fig. \ref{fig:human_approval} shows the human approval prompt during a workflow run.

\begin{figure}[!h]
    \centering
    \includegraphics[width=\columnwidth]{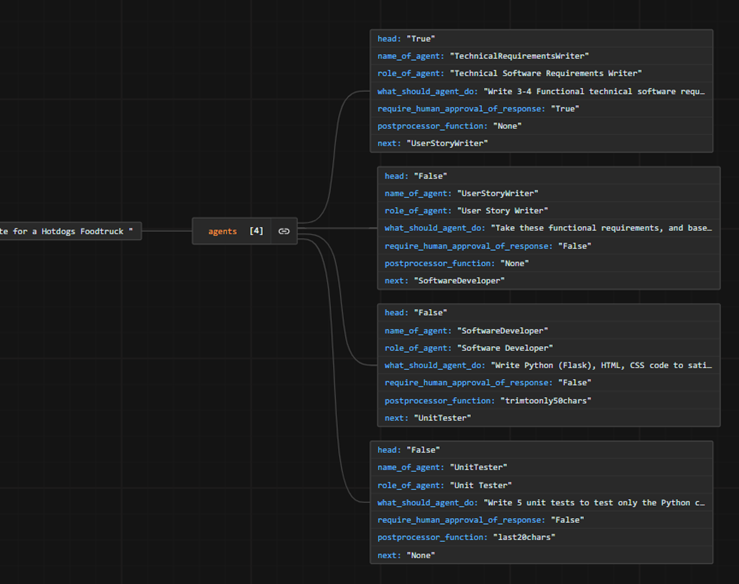}
    \caption{Visualization of the \texttt{foodtruck-website.json} workflow, showing a four-agent sequence for software development.}
    \label{fig:foodtruck_flow_viz}
\end{figure}

\begin{figure}[!h]
    \centering
    \includegraphics[width=\columnwidth]{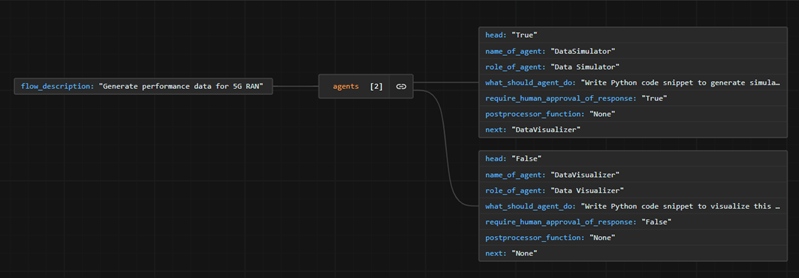}
    \caption{Visualization of the \texttt{5G-RAN-data.json} workflow, a two-agent sequence for data simulation and visualization.}
    \label{fig:5g_flow_viz}
\end{figure}

\begin{figure}[!h]
    \centering
    \includegraphics[width=\columnwidth]{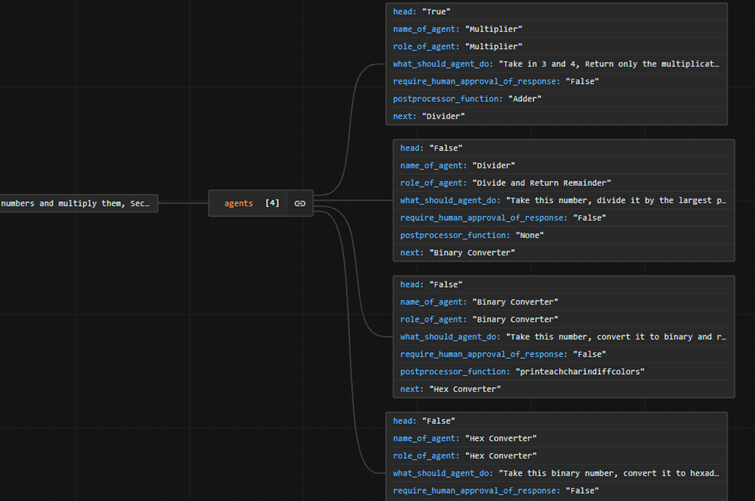}
    \caption{Visualization of the \texttt{mathematician.json} workflow, a four-step sequence involving arithmetic and data conversion agents.}
    \label{fig:math_flow_viz}
\end{figure}

\begin{figure}[!h]
    \centering
    \includegraphics[width=\columnwidth]{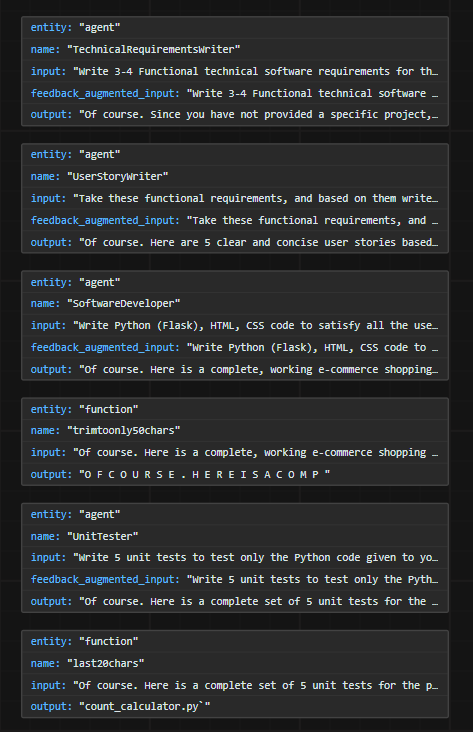}
    \caption{Visualization of the logged interactions for the \texttt{foodtruck-website.json} run, showing the inputs and outputs for each agent and function.}
    \label{fig:foodtruck_interaction_viz}
\end{figure}

\begin{figure}[!t]
    \centering
    \includegraphics[width=0.5\textwidth]{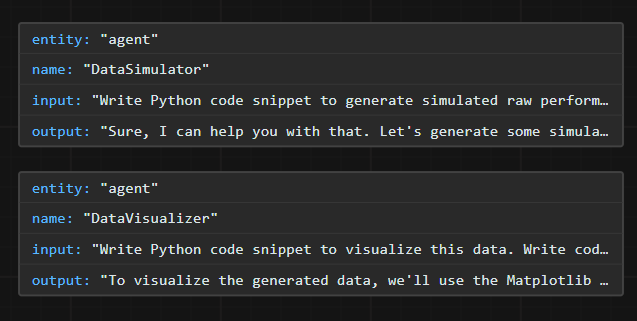}
    \caption{Visualization of the logged interactions for the \texttt{5G-RAN-data.json} workflow.}
    \label{fig:5g_interaction_viz}
\end{figure}

After the flow execution finishes, the individual agent outputs i.e. intermediate outputs as well as the final result of the flow can be obtained as outputs of the function call to execute the flow. They, as well as the agents, preprocessor function outputs are logged as interactions under the 'Interactions' directory as JSONs and hence can be visualized easily. Here we visualize them using an In-IDE visualizer JSON Crack \cite{jsoncrack_github}. Fig. \ref{fig:foodtruck_flow_viz}, Fig. \ref{fig:5g_flow_viz}, and Fig. \ref{fig:math_flow_viz} show visualizations of different workflow structures. Fig. \ref{fig:foodtruck_interaction_viz} and Fig. \ref{fig:5g_interaction_viz} show the corresponding logged interactions for two of these workflows.

\begin{figure}[!t]
    \centering
    \includegraphics[width=0.6\textwidth]{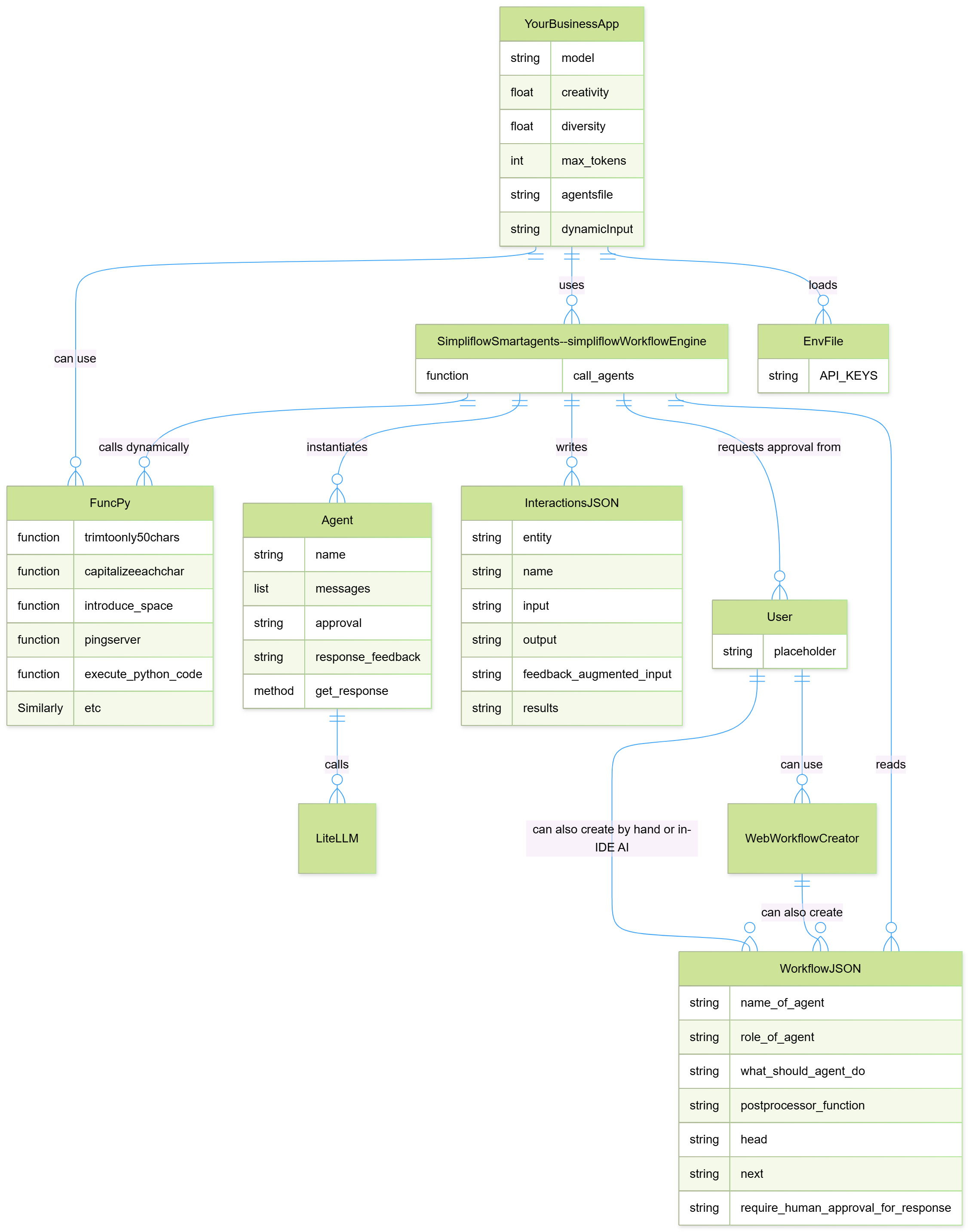}
    \caption{Entity-Relationship (E-R) Diagram for simpliflow. This illustrates the core entities (like YourBusinessApp, Agent, WorkflowJSON) and their relationships, showing how they interact within the framework's ecosystem.}
    \label{fig:er_diagram}
\end{figure}

\begin{figure*}[!t]
    \centering
    \includegraphics[width=\textwidth]{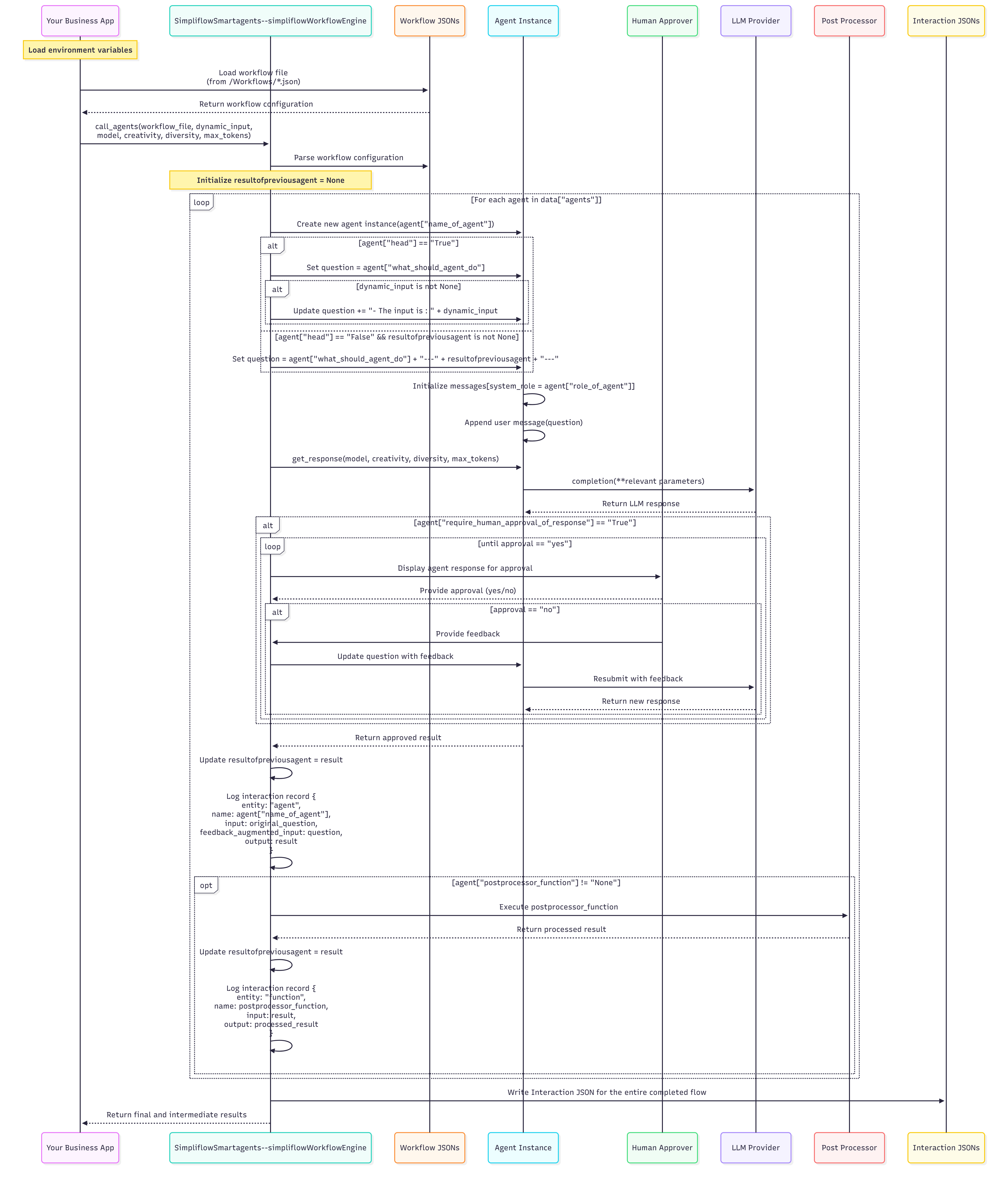}
    \caption{Sequence Diagram for a typical simpliflow workflow execution. It details the step-by-step interactions between the user's application, the workflow engine, the LLM provider, and optional components like the Human Approver and Post Processor.}
    \label{fig:sequence_diagram}
\end{figure*}

\begin{figure*}[!t]
    \centering
    \includegraphics[width=0.65\textwidth]{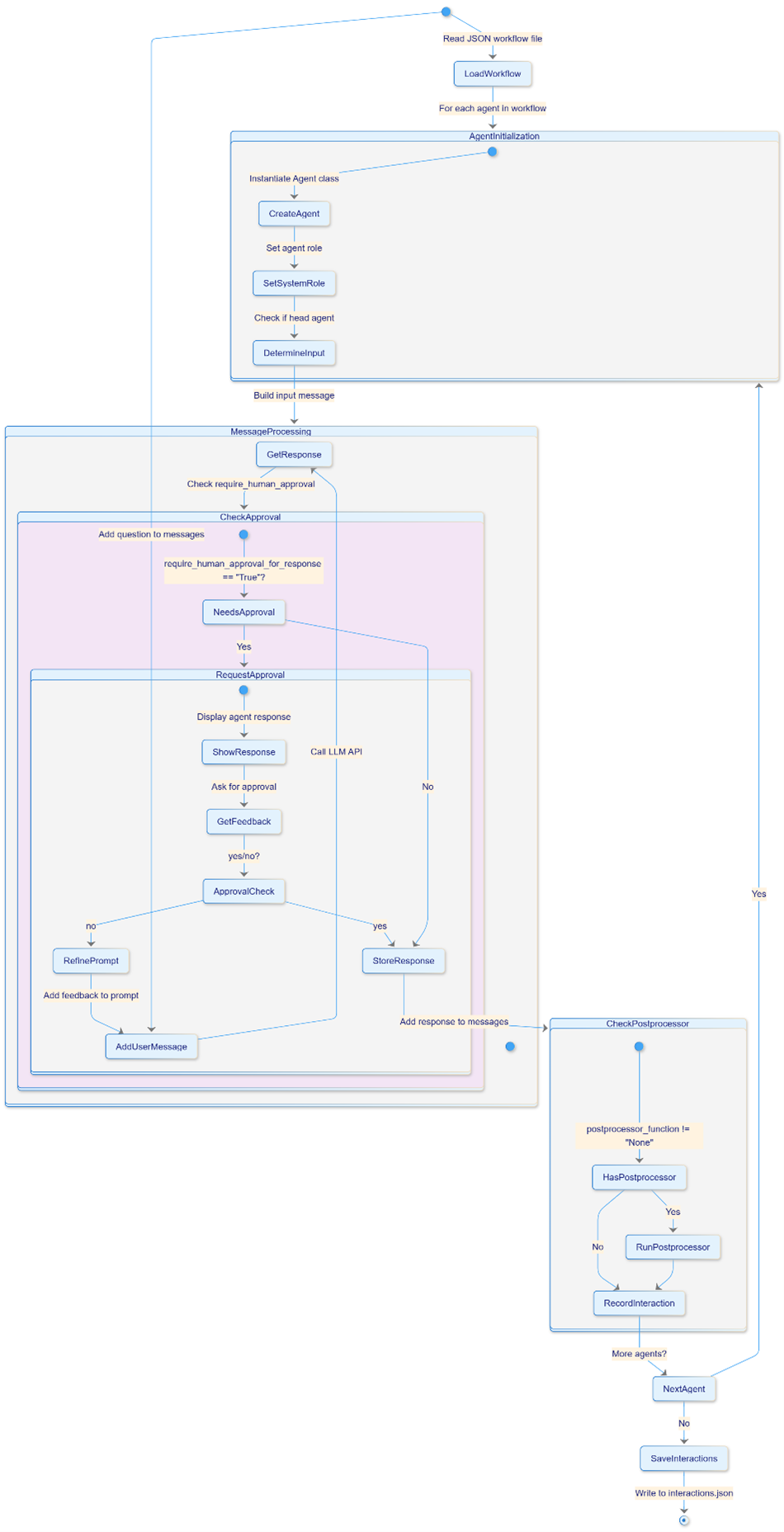}
    \caption{State Diagram for the simpliflow workflow engine.}
        \label{fig:state_diagram}
\end{figure*}

The detailed E-R, Sequence and State diagrams in Fig. \ref{fig:er_diagram}, Fig. \ref{fig:sequence_diagram}, and Fig. \ref{fig:state_diagram} respectively offer a deeper understanding of the working of simpliflow.

\onecolumn

\twocolumn

\FloatBarrier
\section{Features and Capabilities}
The features and capabilities of simpliflow are summarized below -
\subsection{Ease of Use}
\begin{itemize}
    \item Start creating AI Agents as easy as 1-2-3
    \item Can support Infinitely Long Linear workflows – containing any number of Agents and user defined postprocessor functions
    \item Your creativity is the limit
    \item Low barrier to Adoption – Quick and Easy Integration with existing Projects and Workflows
    \item 1-click or No-click Automation
    \item Comes ready with Example flows from various domains
    \item More flows can be created quickly using In-IDE AI's or Interactive Workflow Generator web utility or manually
    \item Power and Control to Developers
    \item Supports Human Approval/ Human-in-the-Loop (HITL)
\end{itemize}

\subsection{Support}
\begin{itemize}
    \item Open-Source
    \item Works with Python \textgreater 3.8
   \item Supports 100+ LLM Vendors and Models through LiteLLM \cite{litellm_github} like Anthropic Claude, Google Gemini, OpenAI ChatGPT, Deepseek, HuggingFace etc.
    \item Supports Locally deployed/operationalized models
    \item Supports multiple use cases/Projects/Workflows
    \item Supports multiple environments (Windows/Linux) and IDEs (VSCode, Spyder, PyCharm or terminal)
    \item Supports multiple Compute (Local, Cloud VMs, etc.)
    \item Supports Tuning of most common LLM Hyperparameters
    \item Makes use of local compute and your existing compute for non-LLM processing
\end{itemize}

\subsection{Debugging}
\begin{itemize}
    \item Extract Final Results or Results of Intermediate agents too.
    \item Results viewable in Interaction files
    \item Visualize both agents and interactions using optional in-IDE JSON plugins like JSON Crack \cite{jsoncrack_github} or other JSON formatters
\end{itemize}

\subsection{Support for User defined Postprocessors}
\begin{itemize}
   \item User defined postprocessor functions can have many other user-defined functions within them - Function Interleaving
    \item The functions can be response formatters, validators, etc.
\end{itemize}

\subsection{Powerful AI to Action}
\begin{itemize}
 \item Code Injection into Execution Environments via Post Processor functions
 \item Post Processor functions allow Code Execution in environments, interpreters, runtime engines, utilities etc. thereby enabling AI to Action capabilities.
\end{itemize}

\subsection{Extensibility and Integration with Other Systems}
\begin{itemize}
   
    \item Flows receive inputs in 3 ways - 1. as an Input to flow (these inputs could be from databases, from RAG, from API's, etc.), 2. Hardcoded in the Workflow JSON, 3. Simulate/Generate or Dynamic Input from Preprocessing, 4. Provide during HITL
    \item Modular nature supports integration with other Python and non-Python systems, other paradigms like RAG, etc.
    (RAG not needed in many cases due to larger token limit of current LLMs but RAG possible with Vector databases installed in the environment – can support multiple Vector DBs, multiple Embedding Models, multiple Chunking Mechanisms)
    \item Integrate with other database systems (ServiceNow, Snowflake, Palantir etc.)
\end{itemize}
   
\subsection{Scalability}
\begin{itemize}
\item Scales easily
\item Execute Multiple different flows using a single line of code.
\item Ability to schedule/triggers for various times and conditions.
\end{itemize}

\section{Discussion and Comparison with Existing Frameworks}
Several frameworks and toolkits have emerged to help developers build AI agent systems. Here we compare simpliflow with several prominent ones – LangChain \cite{langchain_github}, AutoGen \cite{autogen_github}, BabyAGI \cite{babyagi_github}, CrewAI \cite{crewai_github}, and SuperAGI \cite{superagi_github}—focusing on architectural and functional differences. A detailed comparison is provided in Table \ref{tab:comparison}.

\subsection{Extremely Easy Setup and Minimal Boilerplate}
Frameworks such as AutoGen \cite{autogen_github}, LangChain \cite{langchain_github}, and BabyAGI \cite{babyagi_github} are powerful but typically require more setup time and expertise. Simpliflow addresses this by trading some complexity for simplicity and rapid prototyping. In several agent frameworks (e.g., AutoGen), multiple chat semantics can be confusing. Simpliflow aims to remain simple and abstracts much of the complexity away from developers with a single template.

\subsection{Declarative JSON-Based Workflow Definition}
simpliflow's workflow is specified in JSON, which is both human-readable and machine-parsable. The JSON interface is easily extensible to other interaction modalities (e.g., form-based UIs, web editors, or speech-driven builders).

\subsection{Modular Architecture}
Simpliflow decouples agent management, workflow orchestration, and post-processing. The separation of concerns—JSON workflow definition, trigger/execute logic, and user-defined functions (UDFs)—enables workflow construction via external utilities. This clean separation supports independent development and evolution of configuration, business logic, and UDF development. This architecture facilitates easy integration of retrieval-augmented generation (RAG).

\subsection{Control and Determinism}
Execution is governed by an explicit workflow definition, tunable LLM parameters, and transparent result visualization, yielding predictable runs. Simpliflow targets linear, deterministic finite-state workflows; for each input event, there is a single, predictable transition. This design choice reflects the fact that most real-world workflows are linear.

\subsection{Flexible, Dynamic Post-Processing with Python Functions}
Developers can inject custom logic and data transformations at any step. Whereas some frameworks expose many callable functions to an embedded interpreter, Simpliflow routes post-step processing through a single declared postprocessor function that can itself call additional helper functions. As with other modern frameworks, code execution, shells, Docker, CLI, and Jupyter-style utilities can be invoked cleanly via post-processor functions.

\subsection{Execution Semantics and State}
Simpliflow maintains agent state only for the duration of the agent firing. A Simpliflow run completes by executing each agent exactly once in the order defined. With human-in-the-loop enabled, any agent requiring approval is re-executed until approved. It does not rely on caching to repeat identical responses.

\subsection{Output Visibility and Logging}
Unlike some systems where some agent outputs can be silenced, Simpliflow surfaces all agent outputs and logs them as JSON files under an Interactions directory to support retrieval, auditing, and analysis.

\FloatBarrier

\section{Use Cases/Applications}
As explained earlier, simpliflow can help you create and deploy AI agents that can perform diverse tasks across domains. It comes with a collection of example workflows demonstrating its applicability across diverse domains. Below are a few examples included with simpliflow (all available in the simpliflow-usage repository \cite{simpliflow_usage_github}):

\begin{enumerate}
    \item \textbf{Customer Service}: The sentiment analysis workflow can be applied to customer care data to monitor service quality, automate feedback processing, and drive real-time operational improvements. Example included workflow: \texttt{Customer-Care-Sentiment-Analysis.json}

    \begin{mdframed}[backgroundcolor=light-gray, roundcorner=10pt,leftmargin=1, rightmargin=1, innerleftmargin=2, innertopmargin=5,innerbottommargin=5, outerlinewidth=1, linecolor=light-gray]
\begin{lstlisting}[basicstyle=\scriptsize\ttfamily, breaklines=true]
{
    "flow_description": "Perform sentiment analysis on Customer Care Data of a Telecom Company",
    "agents": [
        {
            "head": "True",
            "name_of_agent": "DataSimulator",
            "role_of_agent": "Data Simulator",
            "what_should_agent_do": "Write Python code snippet to simulate raw social media data for a telecom company's customer care service. Generate data about customer queries, responses, timestamps, and user ratings. The generated data should be stored in a Pandas dataframe.",
            "require_human_approval_of_response": "True",
            "postprocessor_function": "None",
            "next": "DataCleaner"
        },
        {
            "head": "False",
            "name_of_agent": "DataCleaner",
            "role_of_agent": "Data Cleaner",
            "what_should_agent_do": "Write Python code snippet to clean the collected social media data. This includes removing duplicates, handling missing values, and normalizing text (e.g., lowercasing, removing special characters). The cleaned data should be stored in a Pandas dataframe.",
            "require_human_approval_of_response?": "False",
            "postprocessor_function": "None",
            "next": "SentimentAnalyzer"
        },
        {
            "head": "False",
            "name_of_agent": "SentimentAnalyzer",
            "role_of_agent": "Sentiment Analyzer",
            "what_should_agent_do": "Write Python code snippet to perform sentiment analysis on the cleaned social media data. Use a pre-trained sentiment analysis model . The results should include sentiment scores and labels (positive, negative, neutral) and should be stored in a Pandas dataframe.",
            "require_human_approval_of_response": "True",
            "postprocessor_function": "None",
            "next": "DataVisualizer"
        },
        {
            "head": "False",
            "name_of_agent": "DataVisualizer",
            "role_of_agent": "Data Visualizer",
            "what_should_agent_do": "Write Python code snippet to visualize the sentiment analysis results. Generate a pie chart for sentiment distribution, a bar chart for sentiment over time, and a word cloud for the most frequent words in positive and negative tweets. The charts should have proper legends, titles, and axes names.",
            "require_human_approval_of_response?": "False",
            "postprocessor_function": "None",
            "next": "None"
        }
    ]
}

\end{lstlisting}
\end{mdframed} 
    
    \item \textbf{Content writing and Brand Management}: The markdown formatter workflow can be expanded for content writing, communications and brand management applications. Example included workflow: \texttt{Dynamic-Input-Example-Apple.json}

    \begin{mdframed}[backgroundcolor=light-gray, roundcorner=10pt,leftmargin=1, rightmargin=1, innerleftmargin=2, innertopmargin=5,innerbottommargin=5, outerlinewidth=1, linecolor=light-gray]
\begin{lstlisting}[basicstyle=\scriptsize\ttfamily, breaklines=true]
{
    "flow_description": "Format the piece of text as Markdown",
    "agents": [
        {
            "head": "True",
            "name_of_agent": "MarkdownFormatter",
            "role_of_agent": "Markdown Formatter",
            "what_should_agent_do": "Take in a piece of text and format it as a multi-line Markdown. Return just the Markdown formatted text. No other extra text",
            "require_human_approval_of_response": "False",
            "postprocessor_function": "None",
            "next": "None"
            }
      
       
    ]
  }


\end{lstlisting}
\end{mdframed} 
    In Your-Business-App.py, you can create/use custom objects and pass them to your flows.

    \begin{mdframed}[backgroundcolor=light-gray, roundcorner=10pt,leftmargin=1, rightmargin=1, innerleftmargin=5, innertopmargin=5,innerbottommargin=5, outerlinewidth=1, linecolor=light-gray]
\begin{lstlisting}[basicstyle=\scriptsize\ttfamily, breaklines=true]
# STEP 1: REQUIRED PARAMETERS
. . . . 
# STEP 2: OPTIONAL PARAMETERS 
. . . .
# STEP 3: SOME MORE CONFIGURATION (agentsfile and dynamic_input)
agentsfile = os.path.join("Workflows", "Dynamic-Input-Example-Apple.json")

# Here is an example of a custom object
class Fruit:
    def __init__(self, name, color, size, price, linktobuy):    
        self.name = name
        self.color = color
        self.size = size
        self.price = price
        self.linktobuy = linktobuy

    def __repr__(self):
        return f"Fruit({self.name}, {self.color}, {self.size}, {self.price}, {self.linktobuy})"

    def __str__(self):
        return f"Fruit Description Fruit A {self.color}, {self.size}, {self.name}. Price: {self.price}. Link to buy: {self.linktobuy}"
        
apple = Fruit("apple", "red", "large", "$3.00/lb", "https://usapple.org/")

dynamic_input = apple
# Always convert dynamic input to string
dynamic_input = str(dynamic_input)  # or repr(dynamic_input)  for your custom objects/classes.

# STEP 4: RUN THE FLOW!
a, b = sim.call_agents(agentsfile, dynamic_input, model, creativity, diversity, max_tokens)


\end{lstlisting}
\end{mdframed} 
    \item \textbf{Network Performance Simulations}: We can simulate or use network performance data, and analyze it quickly to help diagnose issues, and optimize resource allocation across network nodes. Example included workflow: \texttt{5G-RAN-data.json}

\begin{mdframed}[backgroundcolor=light-gray, roundcorner=10pt,leftmargin=1, rightmargin=1, innerleftmargin=2, innertopmargin=5,innerbottommargin=5, outerlinewidth=1, linecolor=light-gray]
\begin{lstlisting}[basicstyle=\scriptsize\ttfamily, breaklines=true]
{
    "flow_description": "Generate performance data for 5G RAN",
    "agents": [
        {
            "head": "True",
            "name_of_agent": "DataSimulator",
            "role_of_agent": "Data Simulator",
            "what_should_agent_do": "Write Python code snippet to generate simulated raw performance data 5G RAN. Generate data about latency, throuhput, users attached, mobility, type of devices, sessions end and start times etc. The generated data should be stored in a Pandas dataframe",
            "require_human_approval_of_response": "True",
            "postprocessor_function": "None",
            "next": "DataVisualizer"
        },
        {
            "head": "False",
            "name_of_agent": "DataVisualizer",
            "role_of_agent": "Data Visualizer",
            "what_should_agent_do": "Write Python code snippet to visualize this data. Write code to generate a barchart and a scattercharts and a donut chart, and a time series chart. The charts should have proper legends, titles, and axes names",
            "require_human_approval_of_response": "False",
            "postprocessor_function": "None",
            "next": "None"
        }
    ]
}



\end{lstlisting}
\end{mdframed} 
    
    \item \textbf{Operational Reporting for Sales}: Automated report generation workflows can compile sales and performance metrics for taking actions. Example included workflow: \texttt{Ecommerce.json}

\begin{mdframed}[backgroundcolor=light-gray, roundcorner=10pt,leftmargin=1, rightmargin=1, innerleftmargin=2, innertopmargin=5,innerbottommargin=5, outerlinewidth=1, linecolor=light-gray]
\begin{lstlisting}[basicstyle=\scriptsize\ttfamily, breaklines=true]
{
    "flow_description": "Simulated raw sales data for an Ecommerce website",
    "agents": [
        {
            "head": "True",
            "name_of_agent": "DataSimulator",
            "role_of_agent": "Data Simulator",
            "what_should_agent_do": "Write Python code snippet to generate simulated raw sales for an Ecommerce website. The generated data should be stored in a Pandas dataframe",
            "require_human_approval_of_response": "True",
            "postprocessor_function": "None",
            "next": "DataVisualizer"
        },
        {
            "head": "False",
            "name_of_agent": "DataVisualizer",
            "role_of_agent": "Data Visualizer",
            "what_should_agent_do": "Write Python code snippet to visualize this data. Also generate some Donut charts",
            "require_human_approval_of_response": "True",
            "postprocessor_function": "None",
            "next": "None"
        }
    ]
}

\end{lstlisting}
\end{mdframed} 
    
    \item \textbf{Prototyping and Testing of IT Applications}: Workflows can be used for rapid prototyping, where requirements are converted to code, and tests are automatically generated. The flows perform tasks that an entire Software Development team would perform. Example included workflow: \texttt{foodtruck-website.json}
    
\begin{mdframed}[backgroundcolor=light-gray, roundcorner=10pt,leftmargin=1, rightmargin=1, innerleftmargin=2, innertopmargin=5,innerbottommargin=5, outerlinewidth=1, linecolor=light-gray]
\begin{lstlisting}[basicstyle=\scriptsize\ttfamily, breaklines=true]
{
    "flow_description": "Design a simple website for a Hotdogs Foodtruck ",
    "agents": [
      {
        "head": "True",
        "name_of_agent": "TechnicalRequirementsWriter",
        "role_of_agent": "Technical Software Requirements Writer",
        "what_should_agent_do": "Write 3-4 Functional technical software requirements for the project given to you. ",
        "require_human_approval_of_response": "True",
        "postprocessor_function": "None",
        "next": "UserStoryWriter"
      },
      {
        "head": "False",
        "name_of_agent": "UserStoryWriter",
        "role_of_agent": "User Story Writer",
        "what_should_agent_do": "Take these functional requirements, and based on them write 5 clear user stories. Each user story should be clear, concise and should be written in the format: As a <type of user>, I want <some goal> so that <some reason>. ",
        "require_human_approval_of_response": "False",
        "postprocessor_function": "None",
        "next": "SoftwareDeveloper"
      },
      {
        "head": "False",
        "name_of_agent": "SoftwareDeveloper",
        "role_of_agent": "Software Developer",
        "what_should_agent_do": "Write Python (Flask), HTML, CSS code to satisfy all the user stories given to you. You must write actual and complete working code. At the end generate a folder structure showing the code files and the code in them. ",
        "require_human_approval_of_response": "False",
        "postprocessor_function": "trimtoonly50chars",
        "next": "UnitTester"
      },
      {
        "head": "False",
        "name_of_agent": "UnitTester",
        "role_of_agent": "Unit Tester",
        "what_should_agent_do": "Write 5 unit tests to test only the Python code given to you. You must write actual and complete working tests",
        "require_human_approval_of_response": "False",
        "postprocessor_function": "last20chars",
        "next": "None"
      }
    ]
  }


\end{lstlisting}
\end{mdframed} 

    \item \textbf{Long tedious tasks}: For e.g. Format 10,000 unformatted network Configuration Yang files. Example included workflow: \texttt{XML-formatter.json}
    
\begin{mdframed}[backgroundcolor=light-gray, roundcorner=10pt,leftmargin=1, rightmargin=1, innerleftmargin=2, innertopmargin=5,innerbottommargin=5, outerlinewidth=1, linecolor=light-gray]
\begin{lstlisting}[basicstyle=\scriptsize\ttfamily, breaklines=true]
{
    "flow_description": "Generate an abnormally formatted sample xml file and then pretty print it",
    "agents": [
        {
            "head": "True",
            "name_of_agent": "XMLGenerator",
            "role_of_agent": "XML Generator",
            "what_should_agent_do": "Generate an abnormally formatted sample xml file describing Layer 2 Network elements. Return just the xml. No other extra text",
            "require_human_approval_of_response": "True",
            "postprocessor_function": "printinpink",
            "next": "XML Formatter"
            },
        {
            "head": "False",
            "name_of_agent": "XML Formatter",
            "role_of_agent": "XML Formatter",
            "what_should_agent_do": "Take this abnormally formatted xml file and pretty print it. The pretty printed file should have proper indentation and should be easy to read.Return just the xml. No other extra text",
            "require_human_approval_of_response": "True",
            "postprocessor_function": "printinpink",
            "next": "None"
          }
       
    ]
  }

\end{lstlisting}
\end{mdframed} 
    \item \textbf{Call APIs}: simpliflow can easily be used to craft queries for APIs and call them to interface and interact with other systems. As an example, create and execute quantum programs on a real Quantum Computer. Example included workflow: \texttt{\seqsplit{Simple-Quantum-Circuit-Creator-And-Executor.json} }
    
\begin{mdframed}[backgroundcolor=light-gray, roundcorner=10pt,leftmargin=1, rightmargin=1, innerleftmargin=2, innertopmargin=5,innerbottommargin=5, outerlinewidth=1, linecolor=light-gray]
\begin{lstlisting}[basicstyle=\scriptsize\ttfamily, breaklines=true]
{
   "flow_description": "Write a simple quantum program to execute on IBM Quantum computer",
     "agents": [
         {
         "head": "True",
         "name_of_agent": "QuantumCircuitCreatorandExecutor",
         "role_of_agent": "Quantum Circuit Creator and Executor",
         "what_should_agent_do": "Write a qiskit program to create 2 qubit quantum circuit and observable. Transpile the circuit for the correct backend using transpile from qiskit.compiler and then apply the transpiled circuit layout to the observable before passing to the estimator using observable_isa = observable.apply_layout(layout=qc). Use the qiskit_ibm_runtime and QiskitRuntimeService and my token stored as IBM_API_TOKEN to login. Then execute the circuit on the IBM Quantum computer with backend = service.least_busy(simulator=False). In the code, make sure you surely assign estimator = Estimator(mode=backend). Print the result value as result[0].  Just return the code. Nothing else. Don't even include ```python or ``` at the beginning or end of the code. Refer to https://docs.quantum.ibm.com/guides/hello-world for example working code.", 
         "require_human_approval_of_response": "True",
         "postprocessor_function": "None",
         "next": "Code changer"
         },
         {
         "head": "False",
         "name_of_agent": "Code changer that changes only the estimator",
         "role_of_agent": "Code changer to change the estimator line",
         "what_should_agent_do": "Change only the estimator line in the code to estimator = Estimator(mode=backend). Now return the entire modified code. Nothing else. Don't include ```python or ``` at the beginning or end of the code.",
         "require_human_approval_of_response": "True",
         "postprocessor_function": "None",
         "next": "None"
     }    
     ,
     {
         "head": "False",
         "name_of_agent": "Change the Estimator.run method to have only 1 positional arguments",
         "role_of_agent": "Change the Estimator.run method to have only 1 positional arguments",
         "what_should_agent_do": "Change accurately the Estimator.run method wrap the circuit and observable as a tuple appropriately in a list/PUB. Now return the entire modified code. Nothing else. Don't even include ```python or ``` at the beginning or end of the code.",
         "require_human_approval_of_response": "True",
         "postprocessor_function": "execute_python_code",
         "next": "None"
     }
     ]
 }

\end{lstlisting}
\end{mdframed} 
    \item \textbf{Perform tasks and use External Tools}: These examples show how simpliflow can be interfaced with environments or utilities to actually execute its results. E.g. \texttt{PingServer.json} can ping a remote server and \texttt{Realtime-Action-Beeper.json} can create and play music from your computer.
    
\begin{mdframed}[backgroundcolor=light-gray, roundcorner=10pt,leftmargin=1, rightmargin=1, innerleftmargin=2, innertopmargin=5,innerbottommargin=5, outerlinewidth=1, linecolor=light-gray]
\begin{lstlisting}[basicstyle=\scriptsize\ttfamily, breaklines=true]
{
    "flow_description": "Ping Server and return the status",
    "agents": [
        {
            "head": "True",
            "name_of_agent": "NLP DNS",
            "role_of_agent": "IP or Domain name finder",
            "what_should_agent_do": "I want to check if Linkedin is reachable. Just output the IP address of Linkedin. No other text",
            "require_human_approval_of_response": "False",
            "postprocessor_function": "pingserver",
            "next": "PingServer"

        }
    ]
}

\end{lstlisting}
\end{mdframed}

\begin{mdframed}[backgroundcolor=light-gray, roundcorner=10pt,leftmargin=1, rightmargin=1, innerleftmargin=2, innertopmargin=5,innerbottommargin=5, outerlinewidth=1, linecolor=light-gray]
\begin{lstlisting}[basicstyle=\scriptsize\ttfamily, breaklines=true]
{
    "flow_description": "Create 2 Beeps from my Windows computer speaker and then play a long beep from my Windows computer speaker for 10 seconds",
    "agents": [
        {
            "head": "True",
            "name_of_agent": "BeepCreator",
            "role_of_agent": "Beep Creator",
            "what_should_agent_do": "Write Python 3.11 code snippet to create 2 beeps from my Windows computer speaker. Return only the code snippet. Nothing else. This is important. Do NOT even include backticks ```python or ``` at the beginning or end of the code snippet.",
            "require_human_approval_of_response": "True",
            "postprocessor_function": "execute_python_code",
            "next": "SineWaveCreator"
        },
        {
            "head": "False",
            "name_of_agent": "LongBeepCreator",
            "role_of_agent": "Long Beep Creator",
            "what_should_agent_do": "Write Python 3.11 code snippet to play a long beep from my Windows computer speaker for 10 seconds. Return only the code snippet. Nothing else. This is important. Do NOT even include ```python or ``` at the beginning or end of the code snippet.",
            "require_human_approval_of_response": "False",
            "postprocessor_function": "execute_python_code",
            "next": "None"
        }
    ]
}


\end{lstlisting}
\end{mdframed} 

    \item \textbf{Specialized Knowledge Work}: Workflows could be created to take on the roles of highly specialized knowledge workers that could for e.g. write and optimize SQL queries (\texttt{\seqsplit{SQL-function-writer-and-optimizer.json}}) or act as a Data Analyst (\texttt{Ecommerce.json}).
\end{enumerate}

Simpliflow's value lies in how it fits into your business application. It orchestrates but does not dictate the content of prompts, tools, where it is used, etc. It can become part of your Python-based business application. Its lightweight nature means that it can be easily incorporated into larger systems. For example, a Python web server could use simpliflow within a standard request-response cycle to perform a task based on the incoming input. For non-Python business apps, you might want to wrap simpliflow behind a small service or worker like REST/gRPC API, CLI, serverless function, etc.

\section{Future Work and Conclusion}
simpliflow presents an alternative in the landscape of agentic AI frameworks. By adopting a declarative JSON-based approach to define linear deterministic workflows, it significantly reduces the barrier to entry for developers and enables unprecedented speed in prototyping and deployment. Its modular architecture, model agnosticism via LiteLLM \cite{litellm_github}, and powerful "AI-to-action" capabilities make it a versatile tool for automating a wide array of sequential tasks. Simpliflow is an active project, and several enhancements are planned to expand its capabilities while retaining its core simplicity. Future work will focus on features that improve the robustness, performance, and usability of the framework for, e.g., Enhanced Error Handling, and Async and Parallel Agent Execution.

\bibliographystyle{IEEEtran}
\bibliography{main}

@misc{simpliflow_github,
  author = {Deven Panchal},
  title = {{simpliflow}: A lightweight, open-source Python framework for building and orchestrating linear, deterministic agentic workflows},
  year = {2024},
  publisher = {GitHub},
  journal = {GitHub repository},
  howpublished = {\url{https://github.com/DevenPanchal/simpliflow}}
}

@misc{simpliflow_usage_github,
  author = {Deven Panchal},
  title = {{simpliflow-usage}: Usage examples and workflows for the simpliflow framework},
  year = {2024},
  publisher = {GitHub},
  journal = {GitHub repository},
  howpublished = {\url{https://github.com/DevenPanchal/simpliflow-usage}}
}

@misc{litellm_github,
  author = {BerriAI},
  title = {{LiteLLM}: Call all LLM APIs using the OpenAI format},
  year = {2023},
  publisher = {GitHub},
  journal = {GitHub repository},
  howpublished = {\url{https://github.com/BerriAI/litellm}}
}

@misc{langchain_github,
  author = {LangChain},
  title = {{LangChain}: Build context-aware, reasoning applications},
  year = {2022},
  publisher = {GitHub},
  journal = {GitHub repository},
  howpublished = {\url{https://github.com/langchain-ai/langchain}}
}

@misc{autogen_github,
  author = {Microsoft},
  title = {{AutoGen}: Enable Next-Gen Large Language Model Applications},
  year = {2023},
  publisher = {GitHub},
  journal = {GitHub repository},
  howpublished = {\url{https://github.com/microsoft/autogen}}
}

@misc{babyagi_github,
  author = {Yohei Nakajima},
  title = {{BabyAGI}: An AI-powered task management system},
  year = {2023},
  publisher = {GitHub},
  journal = {GitHub repository},
  howpublished = {\url{https://github.com/yoheinakajima/babyagi}}
}

@misc{crewai_github,
  author = {Joao Moura},
  title = {{CrewAI}: Cutting-edge framework for orchestrating role-playing, autonomous AI agents},
  year = {2023},
  publisher = {GitHub},
  journal = {GitHub repository},
  howpublished = {\url{https://github.com/joaomdmoura/crewAI}}
}

@misc{superagi_github,
  author = {Transformation AI},
  title = {{SuperAGI}: A dev-first open source autonomous AI agent framework},
  year = {2023},
  publisher = {GitHub},
  journal = {GitHub repository},
  howpublished = {\url{https://github.com/Transformation-AI/SuperAGI}}
}

@misc{jsoncrack_github,
  author = {Aykut Kardas},
  title = {{JSON Crack}: Seamlessly visualize your JSON data instantly into graphs},
  year = {2022},
  publisher = {GitHub},
  journal = {GitHub repository},
  howpublished = {\url{https://github.com/AykutKardas/jsoncrack.com}}
}

\onecolumn
\begin{center}
\captionof{table}{Comparison of AI Orchestration Frameworks\label{tab:comparison}}
\end{center}

\begin{scriptsize}
\begin{longtable}{@{}P{2.3cm} P{2.25cm} P{2.25cm} P{2.25cm} P{2.2cm} P{2.25cm} P{2.25cm}@{}}

\toprule
\textbf{Feature / Aspect} & \textbf{Simpliflow (proposed)} & \textbf{LangChain} & \textbf{AutoGen} & \textbf{BabyAGI} & \textbf{CrewAI} & \textbf{SuperAGI} \\
\midrule
\endfirsthead

\toprule
\textbf{\textbf{Feature / Aspect}} & \textbf{Simpliflow (proposed)} & \textbf{LangChain} & \textbf{AutoGen} & \textbf{BabyAGI} & \textbf{CrewAI} & \textbf{SuperAGI} \\
\midrule
\endfirsthead

\multicolumn{7}{c}%
{{\bfseries \tablename\ \thetable{} -- continued from previous page}} \\
\toprule
\textbf{Feature / Aspect} & \textbf{Simpliflow (proposed)} & \textbf{LangChain} & \textbf{AutoGen} & \textbf{BabyAGI} & \textbf{CrewAI} & \textbf{SuperAGI} \\
\midrule
\endhead

\midrule \multicolumn{7}{r}{{Continued on next page}} \\
\endfoot

\bottomrule
\endlastfoot

\textbf{Orchestration Style} & Predefined linear \textbf{workflow} (FSM-like sequence) – fully deterministic flow control by design. & Flexible chains and dynamic \textbf{agents} (LLM-driven decisions); mix of static pipelines and agent loops. & Dynamic \textbf{multi-agent chat} framework; no built-in fixed process (flows emerge from agent interactions). & \textbf{Autonomous loop} generates and executes tasks continuously until stopped (emergent sequence, not predetermined). & Supports both explicit \textbf{Flows} (event-driven steps) and agent \textbf{Crews} (autonomous team), allowing structured or dynamic behavior. & Typically dynamic \textbf{agent planning} with toolkit; user defines goal, agent decides steps (though can script sequences via GUI config). \\
\textbf{Determinism} & \textbf{Yes: }Execution path is fixed by JSON (only LLM output content varies). Promotes reproducibility and predictability. & \textbf{Partial:} Deterministic if using simple Chains; \textbf{Not} if using Agent mode (outcome path can vary run to run). & \textbf{No:} Agents decide next moves in conversation; flow can diverge. Hard to reproduce exact trajectories without forcing a script. & \textbf{No:} Inherently non-deterministic (self-generated tasks differ each run, depending on outcomes and memory). & \textbf{Yes/Partial:} Flows can be deterministic; Crews introduce non-determinism via agent autonomy. Tends toward determinism when using flow control explicitly. & \textbf{No:} Agents often use AI reasoning to choose actions; results may differ each run. Some determinism if using fixed sequence mode in UI, but core design assumes autonomy. \\
\textbf{Primary Abstraction} & \textbf{Workflow JSON} with agents \& transitions; minimal code (just run it via simpliflow API). & \textbf{Library/API} with classes (Chains, Agents, Tools, Memory, etc.); requires writing Python/JS code or YAML configurations. & \textbf{Library/API} with agent classes; define agents and launch chats programmatically (some no-code UI add-on available). & \textbf{Script/Template }– typically implemented as Python script demonstration; not a full library interface (often integrated into others). & \textbf{Library + Platform:} Code library for Flows/Crews and optional control plane UI for enterprise; requires coding to define roles and flows, plus optional configuration in UI. & \textbf{Platform (GUI + YAML)} – offers a web interface and config files to set up agents, with underlying Python framework. Focus on less coding, more configuration. \\
\textbf{Learning Curve} & \textbf{Low:} Very simple syntax and concept (JSON steps). Quick to get running with provided examples. Overall developer-friendly. & \textbf{Moderate–High:} Steep for newcomers – many concepts (prompts, memory, agents, etc.) and rapidly evolving APIs. Good documentation and community, but complexity is inherent to flexibility. & \textbf{Moderate:} Need understanding of async multi-agent paradigms. The concept of agents chatting is intuitive, but mastering advanced features (teaching agents, custom tools) adds complexity. & \textbf{Low (for basic use): }The original code is short and easy to run. High to modify or extend – since not designed as extensible framework, customizing behavior or adding tools requires significant coding. & \textbf{Moderate:} Core ideas (agents = crew members, tasks = flows events) require some learning. The framework is lean but one must write code to utilize it fully. Enterprise features (like deployment, telemetry) add to learning but are optional. & \textbf{Low–}\textbf{Moderate:} The GUI and templates make basic agent setup easy (non-coders can use it). However, fully leveraging it (custom tools, debugging agent decisions) requires understanding the underlying logic. \\
\textbf{Modularity/Extensibility} &  \textbf{High: }Very modular – plug in any LLM via LiteLLM (100+ models), add custom postprocessor functions (for new actions or integrations) easily. Designed to fit into any project (supports many tools, data, etc.). & \textbf{High:} Provides many extension points (custom tools, custom memory, new chain types). Has a large ecosystem of integrations (vector DBs, model APIs). Coupled somewhat to its abstractions (one must conform to LangChain interfaces when extending). & \textbf{High}: Allows custom agents, custom tool functions, and integration of human agents. Primarily extendable through writing Python subclasses or functions. Multi-language support (it has .NET version) extends reach. & \textbf{Low: }Not modular in design – it’s a specific loop structure. Extensions usually involve merging with another framework (e.g., adding LangChain tools) rather than BabyAGI itself providing extension hooks. & \textbf{High:} Built for customization – users can craft unique crew compositions, define low-level agent behaviors, and integrate any model (including local). Also offers plugin tool set (crewai[tools]). Somewhat complex to extend due to need for code, but very flexible in capable hands. & \textbf{High: }Many built-in tool integrations and memory. Allows adding new “skills” or tools via its plugin system. You can swap in different vector DBs, models, etc. Extensible, though one might need to follow the framework’s conventions. \\
\textbf{Memory \& State Management }& \textbf{Minimalist: }No built-in long-term memory between agents beyond passing the immediate output to next. Each run is fresh (unless user manually feeds prior logs in). Relies on external means if persistent memory needed (BYO vector DB for RAG if required). State is primarily the JSON structure and the in-memory variables during execution. & \textbf{Rich: }Provides memory classes for chat history, knowledge graph, etc. Easily add context memory or tracking of state across steps. Also supports caching of LLM calls to reuse results. State management is a key part of LangChain’s value prop (but adds complexity). & \textbf{Conversation history} is inherently the state (messages exchanged). Also supports function calling where intermediate results can be stored. No global memory beyond the chat unless user adds (though one could integrate external memory). Some focus on LLM self-evaluation to update state (teaching agents). & \textbf{Central:} Uses a vector store as memory to remember completed tasks and results, enabling the loop to “remember” what’s done. The task list is state in memory. No concept of ephemeral per-step state except through that persistent memory. & \textbf{Offers both} transient memory (agent context that persists during a crew’s run) and options for persistent memory (delegation to external stores). Telemetry can log state of each agent. State management is explicit: the developer can choose what each agent sees/retains. CrewAI can run agents sequentially or in parallel threads, so synchronization of state is up to flow design. & \textbf{Yes:} Typically integrates with a vector database or other memory to allow agents to store and retrieve information over long tasks. State can include past tool outputs, conversation, etc., managed by the framework. The platform keeps logs and agent state for monitoring. \\
\textbf{Human Oversight} & \textbf{Yes: }Built-in via require\_human\_approval. Human can intercept at specific steps. Otherwise, user is not involved during run (one-shot execution unless paused). & \textbf{Indirect:} No specific “pause for approval” feature; but developers can insert manual steps or use an AgentTool that represents a human. LangChain mostly assumes either fully automated or externally supervised outside the chain. & \textbf{Yes:} AutoGen can include a human as one of the agents (e.g., an interactive console agent) to provide oversight or input in the multi-agent chat. Not a one-click setting, but supported conceptually (human-in-the-loop as an agent). & \textbf{No:} Not in original design – it’s autonomous until stopped. A user could monitor the console and intervene by aborting or editing code, but no formal mechanism for stepwise approval. & Not directly a one-step toggle; however, one can design a Crew with a human “agent” role or require certain steps to output to console for verification. CrewAI is more about full automation, assuming thorough testing rather than live human interjection. & \textbf{Somewhat:} Through the GUI, a user might inspect agent decisions in real-time and could adjust configurations. But once an agent is running, it’s generally autonomous. No explicit pause/approve feature standard, though enterprise usage may involve a human supervisor restarting or adjusting if the agent stalls. \\
\textbf{Notable Strengths} & \textbf{Simplicity \& Determinism:} Very easy to use and integrate; predictable linear logic; quick to adapt to many domains with little overhead. Great for well-defined process automation with LLMs. & \textbf{Extensiveness \& Community:} Huge array of features and integrations; standard choice for complex LLM apps. Can handle complex dialogues, tool chains, and has active support and improvements. & \textbf{Multi-agent Collaboration:} Enables sophisticated interactions between AI agents (and humans) – good for scenarios where agents brainstorm or critique each other. Offers advanced research features (teaching, self-tuning) and supports cutting-edge use cases in agent research. & \textbf{Autonomy \& Creativity:} Maximizes AI autonomy; useful for exploring open-ended problem solving. Minimal setup to witness an agent that self-generates tasks. Inspired many derivatives (e.g., integrated into other frameworks). & \textbf{Speed \& Control:} High-performance execution, independent (no LangChain bloat). Combines structured flows with autonomous agent teams – versatile for simple or complex tasks. Enterprise readiness (monitoring, security) built-in, making it suitable for production pipelines where both AI freedom and oversight are needed. & \textbf{Feature-Complete \& User-Friendly: }Provides a lot out-of-the-box (UI, memory, tools, safety checks). Lower barrier for non-developers to deploy an AI agent solution. Designed for scalability and robustness in production (logging, fail-safes, etc.). \\
\textbf{Notable Trade-offs} & \textbf{Limited Autonomy:} Not suited for scenarios where the AI needs to figure out the process itself or react outside a fixed script. Lacks built-in long-term memory and learning – relies on static design and user iteration to improve. & \textbf{Complex \& Heavy: }Using LangChain for simple tasks can be overkill. The system can become complex with many moving parts; debugging a large LangChain agent can be difficult. Performance can suffer if not carefully managed (overhead of many components). & \textbf{Orchestration Overhead:} Without a fixed process control, managing a large multi-agent workflow might require custom logic. Potentially unpredictable outcomes require careful testing. Also, being relatively new, it’s evolving (Microsoft has even announced a next-gen Agent Framework), so long-term support is something to watch. & \textbf{Lacks Structure/Safety:} Can easily go off-track or produce nonsensical tasks if the LLM errs. Not directly enterprise-ready; needs to be combined with other tools for things like monitoring or tool usage. Essentially a proof-of-concept made practical only when embedded in a larger system. & \textbf{Requires Coding \& Setup:} Despite its power, one still needs to write and orchestrate code to define crews/flows. The enterprise features (AMP) might need infrastructure. If only a simple linear chain is needed, CrewAI could be unnecessarily complex relative to simpliflow. Also, being newer, its community is smaller than LangChain’s (though growing). & \textbf{Complex Under the Hood:} The simplicity for the user is achieved by a complex underlying system. Debugging why a SuperAGI agent did something might be non-trivial. Also, running a persistent agent server might be resource-intensive. There’s some lock-in to how SuperAGI structures problems (less developer control over each decision). \\

\end{longtable}
\end{scriptsize}

\end{document}